\RequirePackage{fix-cm}
\documentclass[smallextended]{svjour3}       % onecolumn (second format)
\smartqed  % flush right qed marks, e.g. at end of proof
\usepackage{graphicx}
\usepackage{amsmath}
\usepackage[numbers]{natbib}
\usepackage{hyperref}
\usepackage{xurl}
\usepackage{caption}
\usepackage{rotating}
\usepackage{lscape}    

\journalname{Applied Intelligence}
\begin{document}

\title{An Efficient Insect Pest Classification Using Multiple Convolutional Neural Network Based Models%\thanks{Grants or other notes
}
% \subtitle{Do you have a subtitle?\\ If so, write it here}

\titlerunning{Insect Pest Classification Using Multiple CNN Based Models}        % if too long for running head

\author{Hieu T. Ung \and
        Huy Q. Ung  \and
        Binh T. Nguyen.
}

%\institute{Hieu T. Ung \at
%              first address \\
%              Tel.: +123-45-678910\\
%              Fax: +123-45-678910\\
%              \email{fauthor@example.com}           %  \\
%%             \emph{Present address:} of F. Author  %  if needed


\institute{Hieu T. Ung \at
              Vietnam National University in Ho Chi Minh City\\
              University of Science, Vietnam   %  \\
           \and
           Huy Q. Ung \at
              Tokyo University of Agriculture and Technology, Japan\\
              \and
              Binh T. Nguyen (Corresponding Author) \at
              Vietnam National University in Ho Chi Minh City\\
              University of Science, Vietnam\\
              \email{ngtbinh@hcmus.edu.vn}
              }

\date{Received: date / Accepted: date}

\maketitle

\begin{abstract}
Accurate insect pest recognition is significant to protect the crop or take the early treatment on the infected yield, and it helps reduce the loss for the agriculture economy. Design an automatic pest recognition system is necessary because manual recognition is slow, time-consuming, and expensive. The Image-based pest classifier using the traditional computer vision method is not efficient due to the complexity. Insect pest classification is a difficult task because of various kinds, scales, shapes, complex backgrounds in the field, and high appearance similarity among insect species. With the rapid development of deep learning technology, the CNN-based method is the best way to develop a fast and accurate insect pest classifier. We present different convolutional neural network-based models in this work, including attention, feature pyramid, and fine-grained models. We evaluate our methods on two public datasets: the large-scale insect pest dataset, the IP102 benchmark dataset, and a smaller dataset, namely D0 in terms of the macro-average precision (MPre), the macro-average recall (MRec), the macro-average F1- score (MF1), the accuracy (Acc), and the geometric mean (GM). The experimental results show that combining these convolutional neural network-based models can better perform than the state-of-the-art methods on these two datasets. For instance, the highest accuracy we obtained on IP102 and D0 is $74.13\%$ and $99.78\%$, respectively, bypassing the corresponding state-of-the-art accuracy: $67.1\%$ (IP102) and $98.8\%$ (D0).  We also publish our codes for contributing to the current research related to the insect pest classification problem. 
\end{abstract}

\section{Introduction}
\label{intro}
Nowadays, agriculture plays an essential role in many developing countries worldwide, especially in Southeast Asia\footnote{\url{https://asean.org/storage/COVID-19-Pandemic-Implications-on-Agriculture-and-Food-Consumption-Final.pdf}}. It reckons for a substantial portion of each country's GDP and employs a vital part of its workforce. Among those countries, Vietnam and Thai Lan become two of the world's largest agricultural exporters. When rice is one of the most well-known agrarian products exported, other kinds of commodities, including coffee, cocoa, maize, fruits, and vegetables, contribute to the region's GDP. For instance, palm oil is one of the leading agricultural products for both two ASEAN countries, Indonesia and Malaysia.

Insect pests are well-known to be the most threatened to crops and agricultural products. Crop areas such as rice and wheat are easily affected by insect pests, causing a heavy loss to the crop owner. For this reason, protecting crops and agricultural products from insect pests becomes a must action in different ASEAN countries for keeping and increasing the volume and the quality of yearly agricultural products exported. Among that, insect identification is needed for early pest forecasting to prevent further crop damaged. Manually identifying insect pests on a large farm by expert human resources is time-consuming and expensive. Nowadays, following the popularity of high-quality image capture devices and the achievements of machine learning in pattern recognition, an automated image-based insect pest recognition system is promising to reduce the labor cost and do this task more efficiently.
There have been multiple difficulties in extracting useful features for insect classification problems using images. It is challenging to derive discriminative features from the insect image for classification since there are many pest species and variants of their size and shapes. Most recent works applied traditional machine learning methods using hand-crafted features such as GIST, HOG, SIFT, and SURF as proposed in~\cite{oliva2001modeling, dalal2005histograms, lowe2004distinctive, bay2006surf} respectively. However, the hand-crafted features lack representation for the large-scale variant shapes of various objects. 

Convolutional Neural Networks (CNN) based features are recently very adaptive to different specific computer vision tasks. With the success of the CNN-based features in many classification tasks, especially on the ImageNet Large Scale Visual Recognition Competition or ILSVRC\cite{russakovsky2015imagenet}, there have been several works using CNN-based features in the insect pest classification problem.
Wu and colleagues ~\cite{Wu_2019_CVPR} had proven CNN-based features could be more efficient than hand-crafted features in this task. Secondly, metamorphosed insects transform by many distinct stages (e.g., egg, larva, pupa, and adult) in their life. Also, they may have significant extensive inter-species similarities. For each species, an effective method has to capture features for representing large-scale variant shapes. There is no previous research addressing this problem on insect recognition. In bird breed or car model classification, this problem is solved using a fine-grained image classification method (\cite{DBLP:journals/corr/abs-1903-06150, Chen_2019_CVPR}). Fine-grained image classification uses the discriminative features extracted from informative regions of the object and encoding them into vectors for classification.

This paper proposes several methods to handle the problems mentioned above in the insect classification problem.
\begin{enumerate}
\item Firstly, we apply a CNN model with an attention mechanism to make the feature extractor focusing on the insects in the input image. The attention mechanism is necessary since the captured images of insects in the crop often include a complex background containing leaf, dust, branch, etc. 
\item Secondly, we utilize a multi-scale CNN model to capture the features of size-variant insects. 
\item Thirdly, we apply a multi-scale learning method for fine-grained image classification to address the high inter-class similarity problem. 
\item Finally, we use the soft-voting ensemble method to combine these models to improve the performance.
\end{enumerate}
The rest of the paper can be organized as follows. We briefly present all related works in the insect classification problem in Section 2 and describe our proposed techniques in Section 3. We compare our methods with previous techniques and show the experimental results and our discussion in Sections 4 and 5. The paper ends with our conclusion and future works.

\section{Related Work}
Recent years have witnessed the notable performance of deep learning methods in different object recognition problems. One of the well-known competitions related to object recognition is the ImageNet Large Scale Visual Recognition Challenge (ILSVRC), where all participants competed with each other for object detection and image classification at a large scale. Among mixed approaches at ILSVRC, convolutional neural networks, including Inception \cite{Szegedy_2015_CVPR} and ResNet \cite{DBLP:journals/corr/HeZR016}, became the winners of the ILSVRC competition. Other works can be found at \cite{Duy_2017,Binh_2018,Hieu_2020}.

For the insect classification task, Cheng et al. ~\cite{cheng2017pest} presented a new approach using deep residual networks to enhance the performance of the crop pest classification problem based on pest images with complex farmland background. The proposed technique could achieve a classification accuracy of 98.67\% for ten classes of crop pest images, better than a plain deep neural network, AlexNet. 
Liu and colleagues ~\cite{liu2016localization} proposed a new approach for localizing pest insect objects in pest images using saliency detection algorithms and then used a deep convolutional neural network classifying agricultural pest insects. The experimental results showed this technique could bypass the previous methods in terms of a mean Accuracy Precision (``mAP''), which was 0.951.
Wang et al. ~\cite{wang2017crop} investigated an insect pest classification problem using deep convolutional neural networks based on crop pest images. They compared the performance of two selected deep neural networks, LeNet-5 and AlexNet, and measure the effects of both convolutional kernels and the number of chosen layers on the final classification accuracy in various experiments.
Thenmozhi and co-workers \cite{thenmozhi2019crop} studied the crop pest classification problem using Convolutional Neural Networks and measured the performance of the proposed techniques and several pre-trained deep learning architectures, including AlexNet, ResNet, GoogLeNet, and VGGNet, on three different datasets (NBAIR, Xie1, and Xie2). The experimental results showed that the proposed technique could outperform other chosen pretrained methods. 

Wu et al. \cite{Wu_2019_CVPR} presented a newly large-scale benchmark dataset for insect pest recognition (IP102). This dataset has more than 75,000 images belonging to 102 categories, where there are about 19,000 annotated images with bounding boxes for object detection. The authors applied hand-crafted (GIST, SIFT, and SURF) and CNN-Based (extracted by AlexNet, VGGNet, GoogleNet, and ResNet) features to measure the corresponding performance of these methods in the dataset IP102. 
Ren and colleagues ~\cite{ren2019feature} proposed the feature reuse residual network (FR-ResNet) for the insect pest recognition problem, combining features from the initial layer of a residual block with the residual signal and stack the feature reuse residual block to create the proposed network. The experimental results on the dataset IP102 illustrated the improved performance compared to the previous methods. 
Liu et al. ~\cite{liu2020deep} also designed a new residual-based block called deep multi-branch fusion residual network (DMF-ResNet) to learn multi-scale representation. It combines basic residual and the bottleneck residual architecture into the residual module with multiple branches. The outputs of these branches can be concatenated and fed into a new module to recalibrate response adaptively and then model the relationship between these branches. Deep multi-branch fusion residual networks had been created by stacking these blocks and applying them for insect pest classification. They measured the performance of the proposed method and compared it with other state-of-the-art approaches. The experimental results illustrated the enhancement of the proposed technique. Other works related can be found in more details at ~\cite{ayan2020crop,nanni2020insect}.

\section{Methodology}
In this section, we present different approaches using retention attention networks (RANs), feature pyramid networks (FPNs), multi-branch and multi-scale attention networks (MMAL-Nets), and the ensemble technique (ET) for improving the performance of the insect pest classification problem. It is worth noting that these methods have specific advantages. For example, the RANs focus on the most crucial region, FPNs efficiently solve small-scale object problems. In addition, MMAL-Nets practically enhance the fine-grained classification problem for recognizing similar objects. Finally, the ensemble technique can help combine different weak classifiers to create more efficient algorithms having better performance.

\subsection{Residual networks}

\begin{figure}[t]
\centering
\captionsetup{justification=centering}
\includegraphics[width=0.45\textwidth]{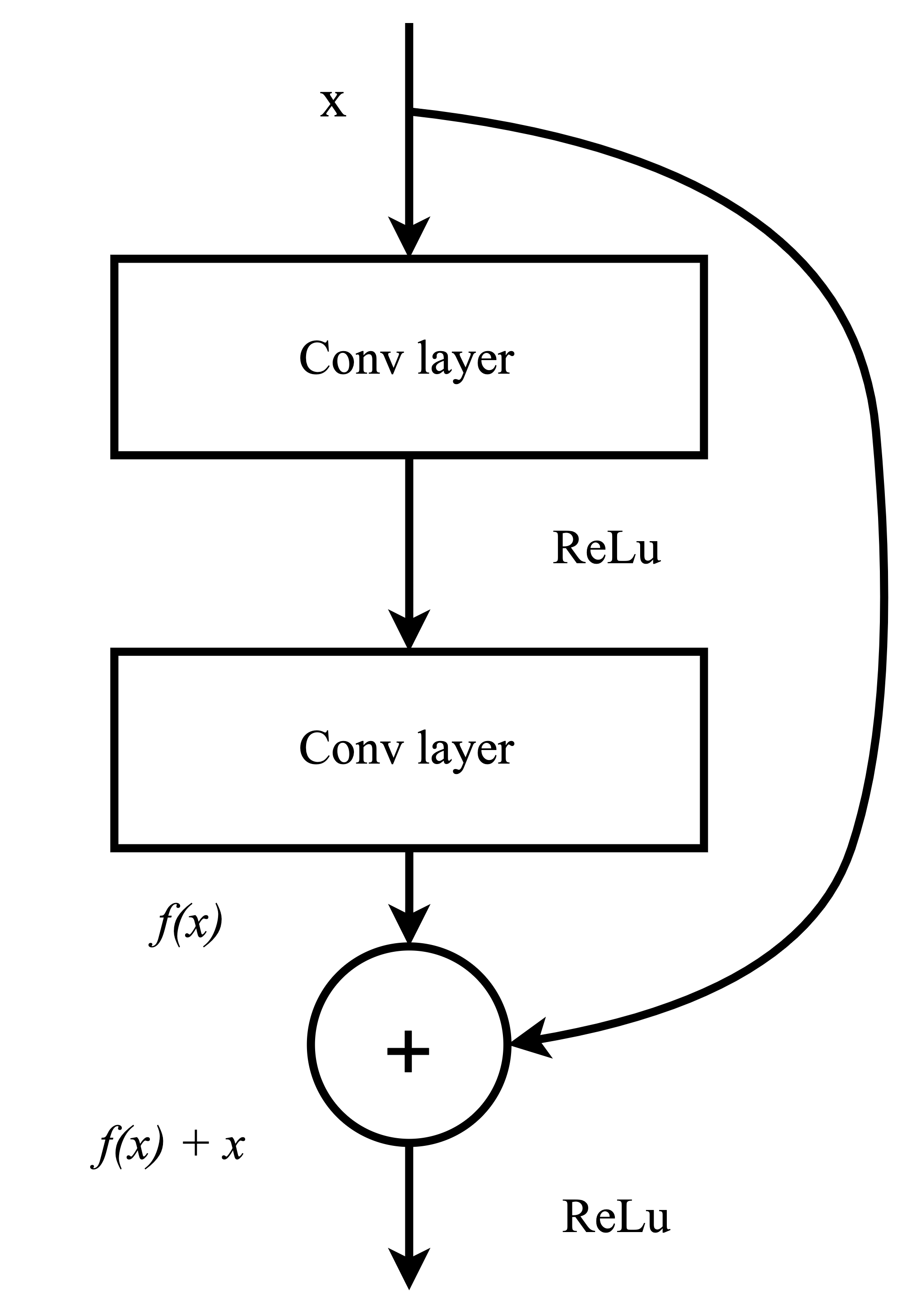}
\caption{The structure of one residual block  in residual networks. Here, $x$ is the input and $f$ is the operation after element-wise addition.}
\label{residual}       % Give a unique label
\end{figure}

Residual networks or ResNet \cite{DBLP:journals/corr/HeZR016} was proposed by He et al. in 2015. The network could help avoid the vanishing gradient problem in training deep neural networks by presenting the skip connection technique among layers. Consequently, the gradient can easily flow back to the input, and the network's weights can be updated. Recently, residual networks can be built by stacking multiple residual blocks (as depicted in Fig.\ref{residual}) to create hugely deep neural networks (maybe up to 1000 layers) depending on each problem. 

\subsection{Residual attention networks}

\begin{figure}
\centering
\captionsetup{justification=centering}
\includegraphics[width=\textwidth]{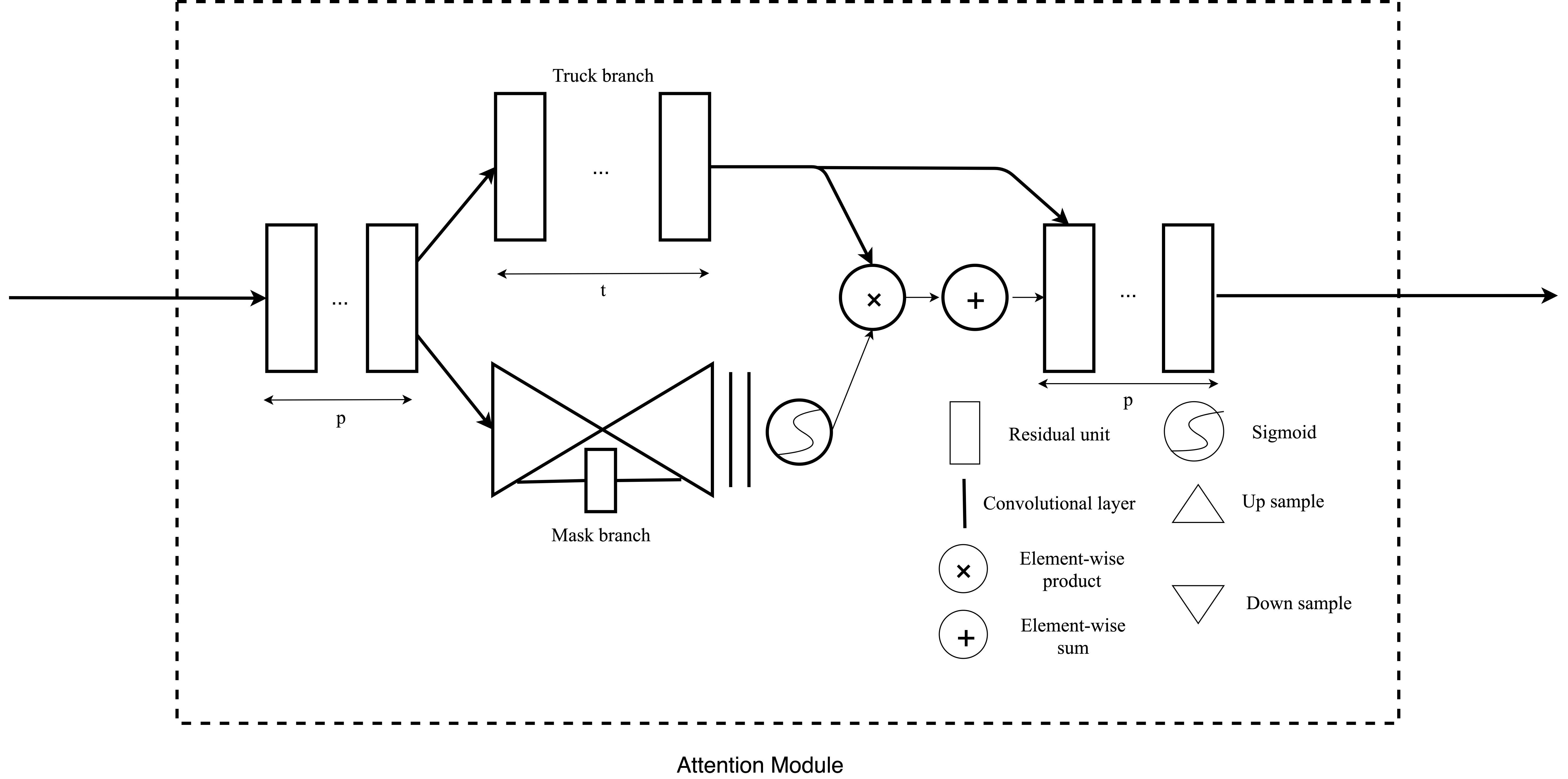}
\caption{The structure of an attention module.}
\label{attres}       % Give a unique label
\end{figure}

Wang and colleagues presented residual attention networks \cite{DBLP:journals/corr/WangJQYLZWT17} (RAN) for image classification problems by adding attention mechanism in CNNs to help the networks to decide which location in the image needs to be focused on. Furthermore, one could build the networks by stacking multiple attention modules that generate attention-aware features to guide the feature learning. As a result, the residual attention networks showed their efficiency at the early stage when surpassing all state-of-the-art image classification methods. 
In general, each attention module can be constructed by two branches: a trunk branch and a mask branch. The ``trunk branch'' performs features extraction, and it can be adapted to any network structure. In this work, we use the pre-activation residual block for the features processing branch. The ``mask branch'' is used for learning attention masks that softly weigh output features. This mask can be used as a feature selector during the feed-forward phase.
Attention module (as depicted in Fig.\ref{attres}) output with residual-attention learning can be formulated by
\begin{equation}
H_{i,c}(x) = (1 + M_{i,c}(x))\times F_{i,c}(x),
\end{equation}
where $x$ is the input, $i$ ranges over all spatial positions, and $c$ is the index of the channel ($c\in{1, ..., C}$). Also, $M(x)$ is the mask branch output and $F(x)$ is the original feature by the trunk branch.

\subsection{Feature pyramid networks}
\begin{figure}[t]
\captionsetup{justification=centering}
\centering
% figures can also be resized by adding to the include graphics command as follows
\includegraphics[width=\textwidth]{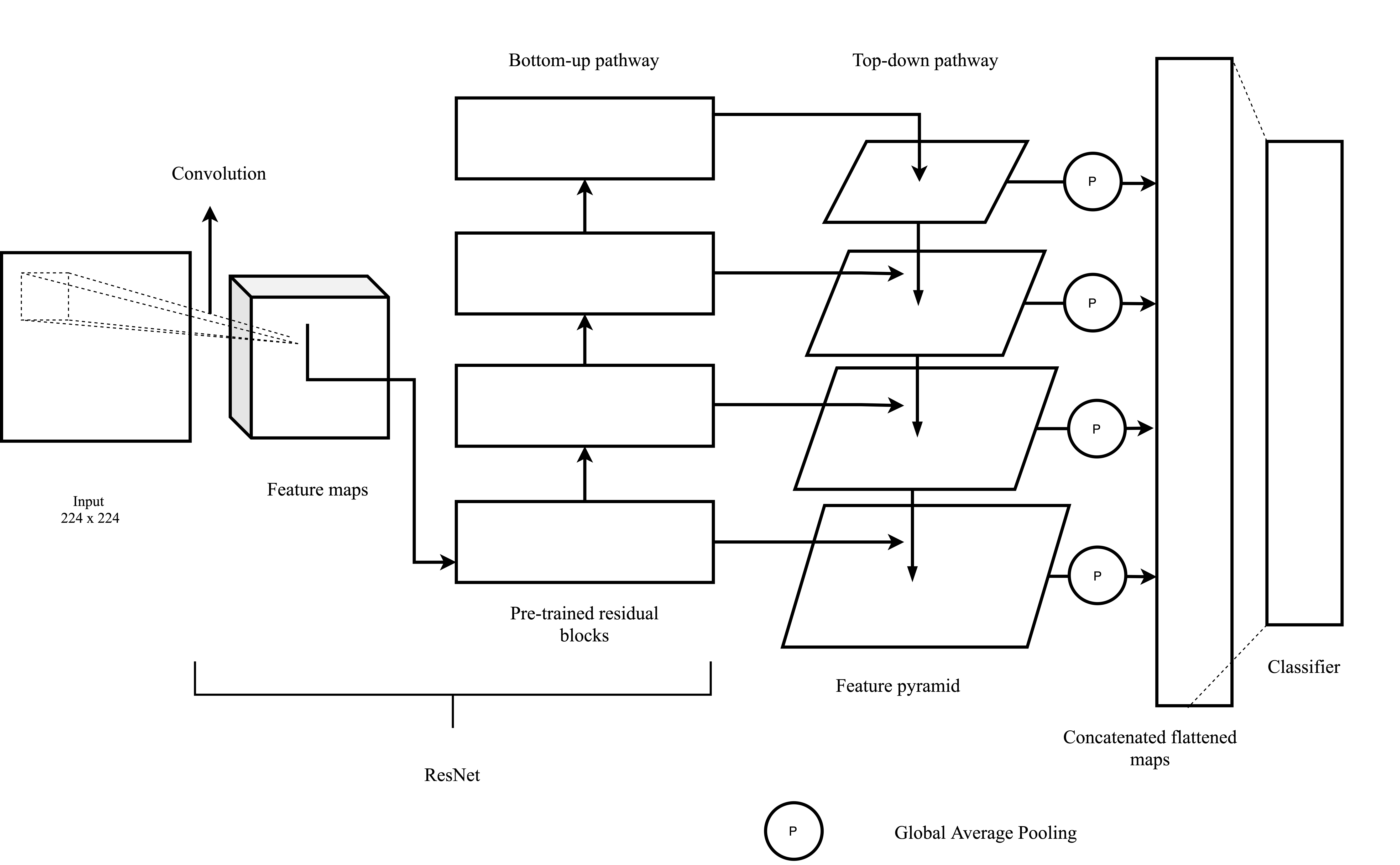}
\caption{Our proposed Feature Pyramid networks' architecture.}
\label{gpns}       % Give a unique label
\end{figure}

\begin{figure}[t]
%\captionsetup{justification=centering}
\centering
% figures can also be resized by adding to the include graphics command as follows
\includegraphics[width=\textwidth]{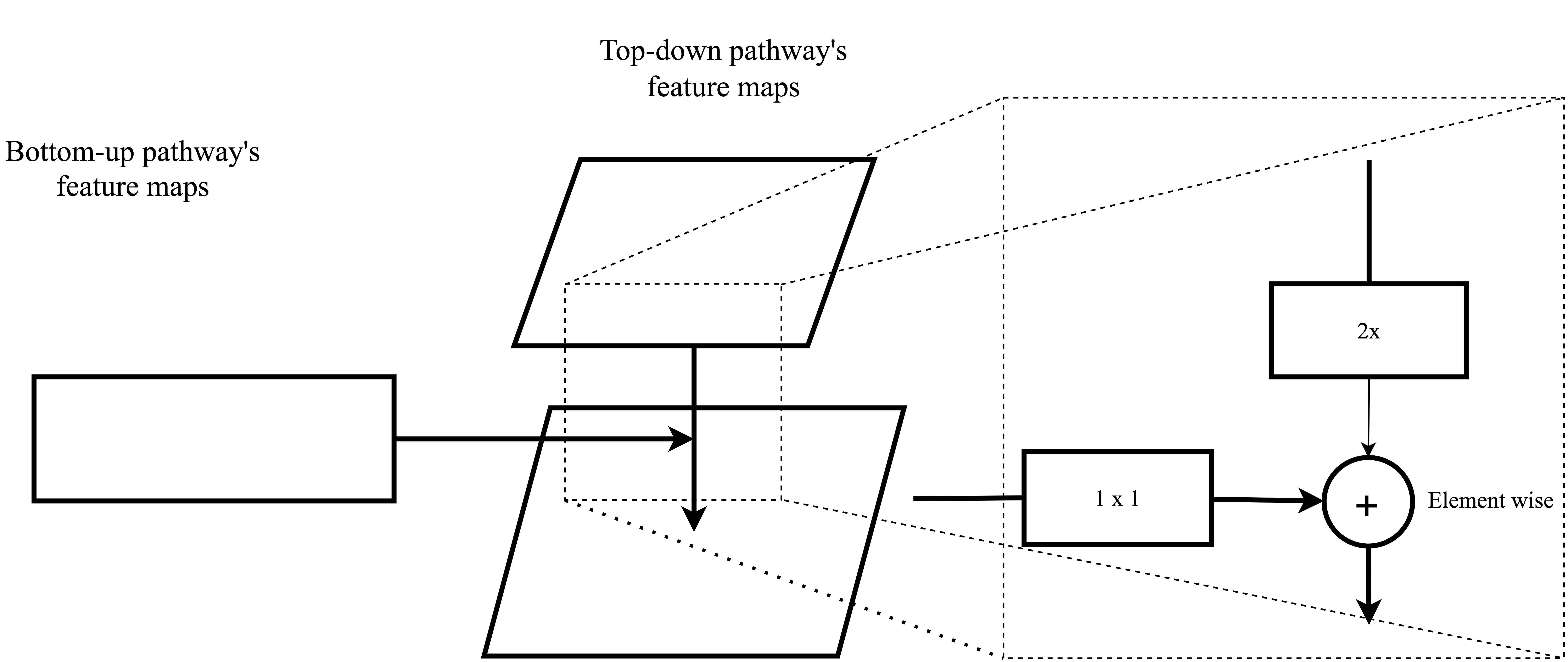}
\caption{The lateral connection in Feature Pyramid Network.}
\label{lateral}       % Give a unique label
\end{figure}

Recognizing objects with highly variant scales is challenging in computer vision. Similarly, in insect classification, the scale of the insect in a captured image is usually small. One way to address this problem is to change the size of a single input image with different scales for constructing a pyramid of multiple input images. However, it makes designed networks bigger and requires much memory and longer training time than using a single input image. Feature Pyramid networks\cite{DBLP:journals/corr/LinDGHHB16} (FPN), is an alternative way to construct a pyramid of features with a small extra cost. FPN is a features extractor with architecture composed of a bottom-up and a top-down pathway. The bottom-up pathway is the usual feed-forward computation of the backbone Convolutional Neural Networks; ResNet is selected in our work. The top-down pathway constructs higher resolution features by up-sampling features maps from higher pyramid levels. These features are then added element-wise with features from the bottom-up pathway via lateral connections (as shown in Fig. \ref{gpns} and Fig. \ref{lateral}). Finally, we apply a $3\times3$ convolution on each merged map to generate the final feature map. This filter reduces the aliasing effect of up-sampling. As a result, final feature maps have the exact spatial sizes and the same numbers of channels.
In our classification model, after generated all pyramid features, we apply global average pooling on each feature map. Then we feed them into the classifier to conduct the final probability distribution result.

\subsection{Multi-branch and multi-scale attention learning network}
Fine-grained image classification is a sub-field of object classification where the classifier distinguishes between visually highly similar objects. The purpose is to make the model focusing on details from coarse level features to fine level features to discriminate similar objects. Many types of research on fine-grained classification methods reached the state-of-the-art on many fine-grained benchmark datasets. In this work, we apply the multi-branch and multi-scale attention learning networks (MMAL-Net) \cite{zhang2020three} for fine-grained image classification on our pest classification task.

\begin{figure}[t]
\centering
%\captionsetup{justification=centering}
% figures can also be resized by adding to the include graphics command as follows
\includegraphics[width=1.0\textwidth]{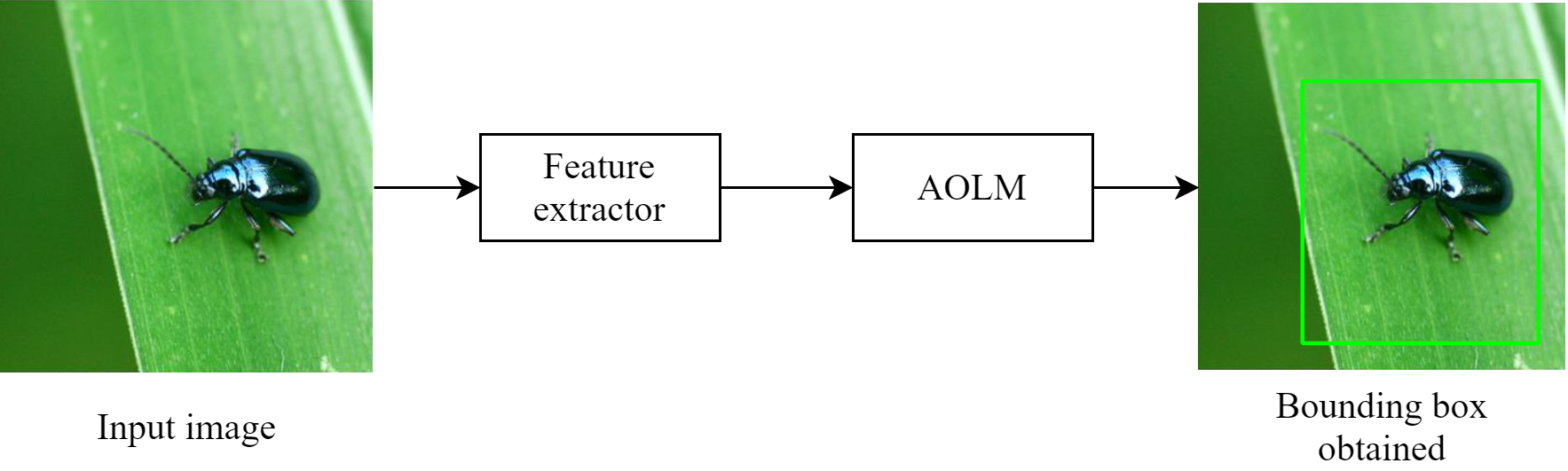}
\caption{AOLM obtained the bounding box using the feature maps from the feature extractor.}
\label{aolm}       % Give a unique label
\end{figure}

The key of the fine-grained classification is to identify informative regions in an image accurately. Usually, we need to localize the object and discriminate parts by drawing bounding boxes by hand. In MMAL-Net, we do not have to do extra annotations, object localization, and multiple discriminate part localization being done automatically by two modules with only the category labels: Attention object location module (AOLM) and attention part proposal (APPM) module. MMAL-Net has three branches in the training phase: a raw branch, an object branch, and a parts branch, all of them using the same ResNet-50 as the features extractor and dense layers as the classifier. In the raw branch, the networks mainly study the overall characteristics of the object. Then the AOLM needs to obtain the object's bounding box information with the help of the features maps of the raw image from its branch (as visualized in Fig.~\ref{aolm}). Thus, the accuracy of object localization is achieved by only using category labels. After obtaining the object's bounding box, we crop the input image following by the bounding box's coordinates to get the finer scale of the object image, and it can be used as the input of the object branch. Finally, the object branch can learn to obtain the final classification result with the input containing structural features and the fine-grained features of the object.

\begin{figure}[t]
\centering
%\captionsetup{justification=centering}
% figures can also be resized by adding to the include graphics command as follows
\includegraphics[width=1.0\textwidth]{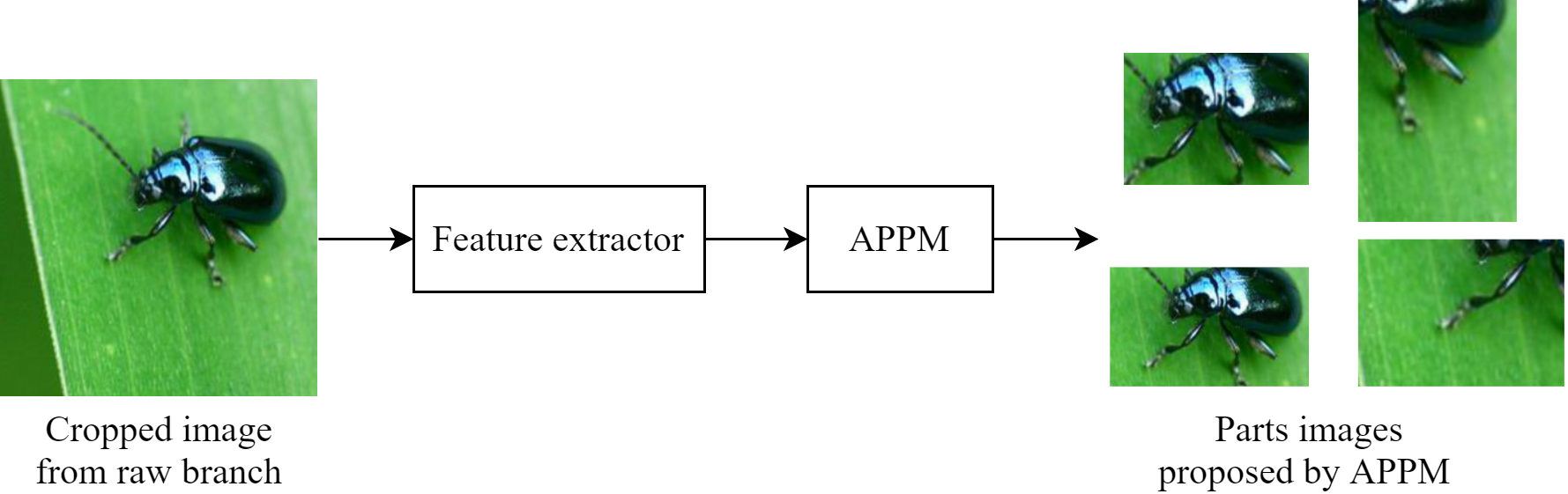}
\caption{APPM proposed the parts image of the object using the feature maps from the feature extractor and the crop image from the raw branch.}
\label{appm}       % Give a unique label
\end{figure}

Additionally, with the feature maps from the object branch, APPM proposes several part regions of the object, which can be used as the input for the parts branch (as shown in Fig.~\ref{appm}). The part images cropped from the object image can train the networks to learn the fine-grained features of different parts in different scales. In the testing phase, the part branch can be disabled, and the final result can be obtained by ensembling the logits from the local branch. 
We combine the logits from the raw branch and local branch to obtain the final result in our work.

\subsection{Ensemble method}
Ensemble learning is a machine learning technique that combines accuracy-low models to obtain more accurate predictions. There are many ways to combine models; in this work, we combine the multiple model's predictions. Combining the models' prediction results can reduce its variance and the generalization error. Specifically, soft voting is one of simple, fast, and reliable ensemble method. The soft-voting method is that we take the sum of all the member models' prediction results on each sample, and then we divide by the number of ensemble members. The final result is the class with the highest probability. Suppose there are $m$ members model and a classification task with $n$ labels. $P_{ij}$ is the predicted probability of model $i = 1, ..., m$ for label $j = 1, ..., n$. One can calculate the ensemble result as follows:
\begin{equation}
P_j = \frac{\sum^{m}_{i = 1}{P_{ij}}}{m},
\end{equation}
where $P_j$ is the predicted probability of class $j$.

\section{Experiments}

This section presents our experiments and compares the proposed approach with the state-of-the-art methods related to the main problem.
We evaluated our proposed models on two datasets, i.e., IP102 and D0, and measure the corresponding performance with the previous methods using standard evaluation metrics, including precision, recall, F1-score, accuracy, and geometric mean score.

\subsection{Datasets}
\begin{figure}[t]
\centering
%\captionsetup{justification=centering}
% figures can also be resized by adding to the include graphics command as follows
\includegraphics[width=\textwidth]{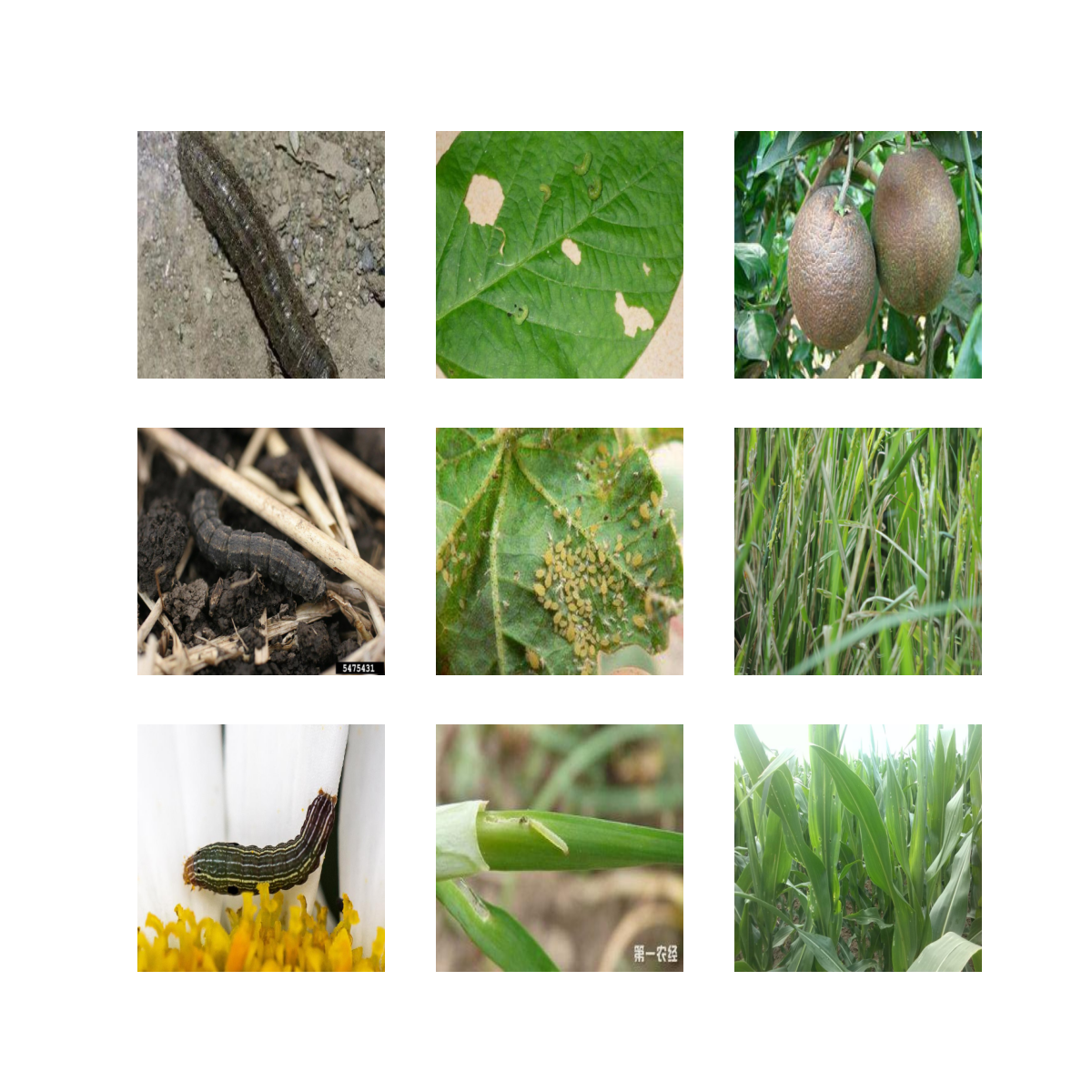}
\caption{Examples for challenges in IP102. Column (a) presents three different worm species, but they are hard to be distinguished. Column (b) shows examples of low-scale insects. Column (c) gives examples of damaged crop fields.}
\label{problem}       % Give a unique label
\end{figure}

We evaluated our proposed method on two datasets. The first dataset is IP102, which is a large-scale benchmark dataset presented in~\cite{Wu_2019_CVPR}. It contains 75.222 images of 102 insect pest species. This dataset has some challenges practically. Firstly, several classes have highly intra-class variances, as shown in Fig.~\ref{problem}(a). Secondly, there are images captured of the damaged crop as shown in Fig. ~\ref{problem}(b). Thirdly, there are images including small-scale insects on the noisy backgrounds as shown in Fig. ~\ref{problem}(c). Finally, there is a significant imbalance among the number of samples in classes.

\begin{figure}[t]
\centering
%\captionsetup{justification=centering}
% figures can also be resized by adding to the include graphics command as follows
\includegraphics[width=0.9\textwidth]{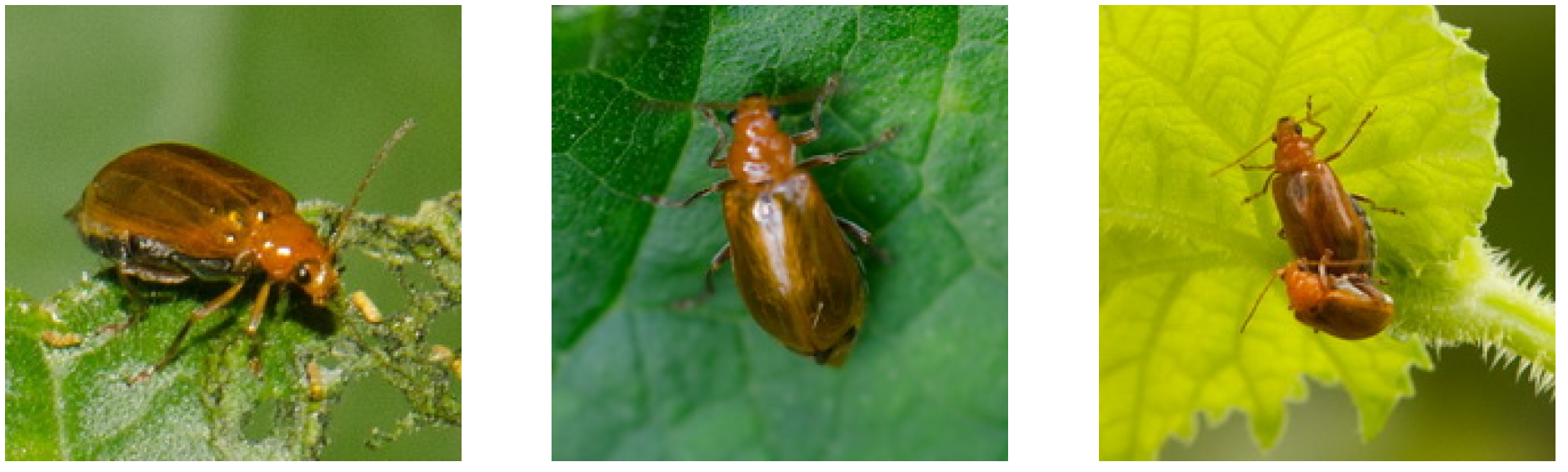}
\caption{Examples of Aulacophora indica gmelin in D0.}
\label{d0image}       % Give a unique label
\end{figure}

The second dataset is D0, presented in~\cite{xie2018multi}. It contains 4.508 images belonging to 40 insect pest species captured in the natural environment. Some examples are shown in Fig.~\ref{d0image}.

\subsection{Experiment settings}
According to~\cite{Wu_2019_CVPR}, IP102 is partitioned into three subsets: a training set of 45.095 images, a validation set of 7.508 images, and a testing set of 22.619 images. We applied this setting in our experiments. For D0, we arbitrarily partitioned it into three subsets, i.e., a training set, a validation set, and a testing set, with a ratio of $7:1:2$.

We applied pre-processing steps on the input image of the size $h\times w$, where h and w are its height and width, respectively. Firstly, we resized the input image into $h'\times w'$ in which the aspect ratio of the original image is kept. First, the smaller value between $h$ and $w$ is resized to 256. Then, the larger value is assigned by multiplying it with the ratio of the bigger value and the smaller value. Secondly, we applied the random crop data augmentation with the window size of $256\times256$ on the training phase to address the over-fitting problem. Finally, we applied the center crop method with the same window size as the training phase in the testing phase.

The settings of all RAN, FPN, and MMAL-Net are set at Table \ref{tab:4}  according to \cite{DBLP:journals/corr/WangJQYLZWT17,DBLP:journals/corr/LinDGHHB16, zhang2020three}, respectively. We used a pre-trained ResNet-50 on ImageNet to initialize trainable weights in FPN and MMLA-Net since they utilize ResNet-50 as the feature extractor. For RAN, we initialized the weights arbitrarily.

In the training phase, we used categorical cross-entropy as the cost function. We utilized the Adam optimizer with the initial learning rate of $10^{-4}$, and the $\beta_1$ and $\beta_2$ coefficients are $0.9$ and $0.999$, respectively. We used exponential decay for scheduling the learning rate with a decay rate of $0.96$. The mini-batch size and the maximum number of training epochs are set as 64 and 100, respectively. The training phase is stopped when the performance on the validation set doesn't improve after ten epochs. We applied the dropout technique with the drop rate of $0.5$ to address the over-fitting problem. To use the ensemble method on RAN, FPN, and MMAL-Net, we trained them in the same training set and tested by voting their predictions.

\begin{table}
\centering
\caption{General training settings for each network}
\label{tab:4}       % Give a unique label
\resizebox{0.81\textwidth}{!}{\begin{minipage}{\textwidth}
\begin{tabular}{lllll}
\hline\noalign{\smallskip}
Model/& ResNet50 & RAN & FPN & MMAL-Net\\
Hyper-Parameters&&&&\\
\noalign{\smallskip}\hline\noalign{\smallskip}
Learning rate &0.0001 & 0.1 & 0.0001 & 0.001\\
Batch size & 64 & 32 & 32 & 6 \\
Optimizer & Adam & SGD & Adam & SGD\\
& $betas=$ & $momen-$ & $betas=$ & $momen$\\
& $(0.9, 0.999)$ & $tum=0.9$ & $(0.9, 0.999)$ & $tum=0.9$\\
Scheduler & Exponential & MultiStep & Exponential & MultiStep\\
& decay rate = 0.96  & decay rate = 0.1 & decay rate = 0.96 & decay rate = 0.1\\
Weight decay & 0.00001 & 0 & 0.00001 & 0.00001 \\
Dropout & 0.3 & 0 & 0 & 0\\
Maximum epochs & 100 & 100 & 100 & 150\\
Input Size & $224\times224$ & $224\times224$
& $224\times224$ & $448\times448$\\
\noalign{\smallskip}\hline
\end{tabular}
\end{minipage}}
\end{table}

\subsection{Evaluation metrics}
We evaluated our proposed models with several suitable metrics for the imbalance among classes in IP102 and D0. The metrics consist of the macro-average precision (MPre), the macro-average recall (MRec), the macro-average F1-score (MF1), the accuracy (Acc), and the geometric mean (GM). To treat the classes equally important, we computed the recall for each category, then took an average of them to obtain \text{MRec} as follows:
\begin{equation}
\text{Rec}_c = \frac{\text{TP}_c}{\text{TP}_c + \text{FN}_c}
\end{equation}

\begin{equation}
\text{MRec} =  \frac{\sum_{c=1}^C{\text{Rec}_c}}{C}
\end{equation}
where \textit{C} is the number of classes. $\text{TP}_c$ and $\text{FN}_c$ stand for the true positive and the false negative of the \textit{c}-th class respectively. Similarly, we computed $\text{Pre}_c$ and $\text{MPre}$ as follows:

\begin{equation}
\text{Pre}_c = \frac{\text{TP}_c}{\text{TP}_c + \text{FP}_c}
\end{equation}

\begin{equation}
\text{MPre} = \frac{\sum_{c=1}^C{\text{Pre}_c}}{C}
\end{equation}
where $\text{FP}_c$ stands for the false positive of the \textit{c}-th class. \text{MF1} is the harmonic mean of \text{MRec} and \text{MPre} as follows:

\begin{equation} \label{eq:7}
\text{MF1} = 2\frac{\text{MPre} \cdot \text{MRec}}{\text{MPre}+\text{MRec}}
\end{equation}

\text{Acc} is computed by the true positive value among all classes as follows:

\begin{equation} \label{eq:8}
\text{Acc} = \frac{\text{TP}}{N}
\end{equation}
where \textit{N} is the number of samples. GM is calculated based on the sensitivity of each class (denoted as $S_c$). $S_c$ and GM are as follows:

\begin{equation}
\text{S}_c = \frac{\text{TP}_c}{\text{TP}_c + \text{FN}_c}
\end{equation}

\begin{equation}
\text{GM} = \prod_{c=1}^{C}{\sqrt[C]{\text{S}_c}}
\end{equation}

GM equals 0 if and only if one of $\text{S}_c$ equals 0. To avoid this problem, we replaced $\text{S}_c$ of 0 by $0.001$.

\subsection{Results}
  We conducted experiments to compare our proposed models and ResNet-50 as a baseline. Table \ref{tab:1} presents the results of those models on IP102. Among those single models, MMAL-Net achieves the best performance on Acc, MPre, MRec, and MF1, which is better at 1.36 percentage points of Acc than ResNet-50. However, the GM score of MMAL-Net is slightly lower than those of FPN and ResNet-50. It implies that the predictions of MMAL-Net on the minor classes are less accurate than those on the major classes. Besides, FPN and ResNet-50 yield comparable results. RAN achieves the lowest results. Combining RAN, FPN, MMAL-Net, and ResNet-50 by the ensemble method performs the best, better 1.98 percentage points than MMAL-Net.
  
  \begin{table}
\centering
\caption{The comparison among different proposed models on IP102.}
\label{tab:1}       % Give a unique label
\begin{tabular}{llllll}
\hline\noalign{\smallskip}
Model/Metric & Acc & MPre & MRec & MF1 & GM  \\
\noalign{\smallskip}\hline\noalign{\smallskip}
ResNet-50 & 70.79 & 62.89 & 65.71 & 63.89 & 59.70 \\
RAN & 62.82 & 55.46 & 57.38 & 56.09 & 51.95 \\
FPN & 70.42 & 62.52 & 64.74 & 63.27 & 59.59 \\
MMAL-Net & 72.15 & 62.63 & 69.13 & 64.53 & 58.43 \\
Ensemble model & \textbf{74.13} & \textbf{65.72} & \textbf{70.74} & \textbf{67.65} & \textbf{62.52} \\
\noalign{\smallskip}\hline
\end{tabular}
\end{table}

Similarly, we conducted experiments to compare between those models on D0. Table~\ref{tab:2} presents the experiment results. Overall, the results on D0 are significantly better than those on IP102. Among the single models, MMAL-Net again yields the best results. FPN and RAN have slightly lower performance than ResNet-50. RAN performs the worst. The ensemble model of RAN, FPN, and MMAL-Net achieves the highest performance.
RAN performed significantly worse than other models on both IP102 and D0. It is probably because the pre-trained ResNet-50 did not initialize trainable weights of RAN on the training phase.
\begin{table}
\centering
\caption{The comparison among different proposed models on D0.}
\label{tab:2}       % Give a unique label
\begin{tabular}{llllll}
\hline\noalign{\smallskip}
Model/Metric & Acc & MPre & MRec & MF1 & GM  \\
\noalign{\smallskip}\hline\noalign{\smallskip}
ResNet-50 & 99.34 & 99.12 & 99.20 & 99.14 & 99.09 \\
RAN & 93.27 & 92.93 & 93.57 & 93.06 & 92.65 \\
FPN & 99.23 & 99.08 & 99.09 & 99.07 & 99.06 \\
MMAL-Net & 99.56 & 99.48 & 99.50 & 99.48 & 99.46 \\
Ensemble model & \textbf{99.78} & \textbf{99.66} & \textbf{99.71} & \textbf{99.68} & \textbf{99.65} \\
\noalign{\smallskip}\hline
\end{tabular}
\end{table}

We compared our proposed method with the previous methods as shown in Table \ref{tab:3}. For IP102, we compared with ResNet-50 implemented in~\cite{Wu_2019_CVPR}, and some variants of ResNet, i.e. FR-ResNet and DMF-ResNet, proposed in~\cite{ren2019feature} and~\cite{liu2020deep}, respectively. The results show that our MMAL-Net outperforms those models. In addition, our ensemble models of RAN, FPN, and MMAL-Net are significantly better than the ensemble methods proposed in~\cite{nanni2020insect} and~\cite{ayan2020crop}. For D0, our proposed models are better than those proposed in~\cite{xie2018multi} and~\cite{ayan2020crop}.

One can see that our implemented ResNet-50 is significantly better than the one implemented in~\cite{Wu_2019_CVPR}. The main difference between the two models is that we applied the random crop augmentation technique and the Adam optimizer while~\cite{Wu_2019_CVPR} did not utilize the augmentation technique and use the Stochastic Gradient Descent optimizer.

\begin{table}
\centering
\caption{The comparison between our proposed models and the previous works. (EM: ensemble method)}
\label{tab:3}       % Give a unique label
\resizebox{0.81\textwidth}{!}{\begin{minipage}{\textwidth}
\begin{tabular}{cllccc}
\hline\noalign{\smallskip}
Dataset& Name & Method & Acc & MF1 & GM \\
\noalign{\smallskip}\hline\noalign{\smallskip}
IP102&\citet{Wu_2019_CVPR} & ResNet-50 & 49.4 & 40.1 & 31.5\\
&\citet{ren2019feature} & FR-ResNet & 55.2 & 54.1 & - \\
&\citet{liu2020deep} & DMF-ResNet & 59.1 & 58.1 & - \\
& \textbf{Ours} & \textbf{MMAL-Net} & \textbf{72.2} & \textbf{64.6} & \textbf{58.4} \\
&\citet{nanni2020insect} & Saliency method & 61.4 & - & - \\
&& + CNNs + EM&&&\\
&\citet{ayan2020crop} & CNNs + EM & 67.1 & 65.8 & - \\
&\textbf{Ours} & \textbf{RAN + FPN }
& \textbf{74.1} & \textbf{67.7} & \textbf{62.5}\\
&&\textbf{+ MMAL-Net + ResNet50}&&&\\
\noalign{\smallskip}\hline\noalign{\smallskip}
D0&\citet{xie2018multi} & MLLF + MKB & 89.3 & - & - \\
& \textbf{Ours} & \textbf{MMAL-Net} & \textbf{99.6} & \textbf{99.5} & \textbf{99.5} \\
&\citet{thenmozhi2019crop} & CNNs & 96.0 & - & - \\
&\citet{ayan2020crop} & CNNs + EM & 98.8 & 98.8 & - \\
&\textbf{Ours}  & \textbf{RAN + FPN} & \textbf{99.8} & \textbf{99.7} & \textbf{99.7} \\
&&\textbf{+ MMAL-Net + ResNet50}&&&\\
\noalign{\smallskip}\hline
\end{tabular}
\end{minipage}}
\end{table}

\begin{figure}[t]
\centering
\includegraphics[width=1.0\textwidth]{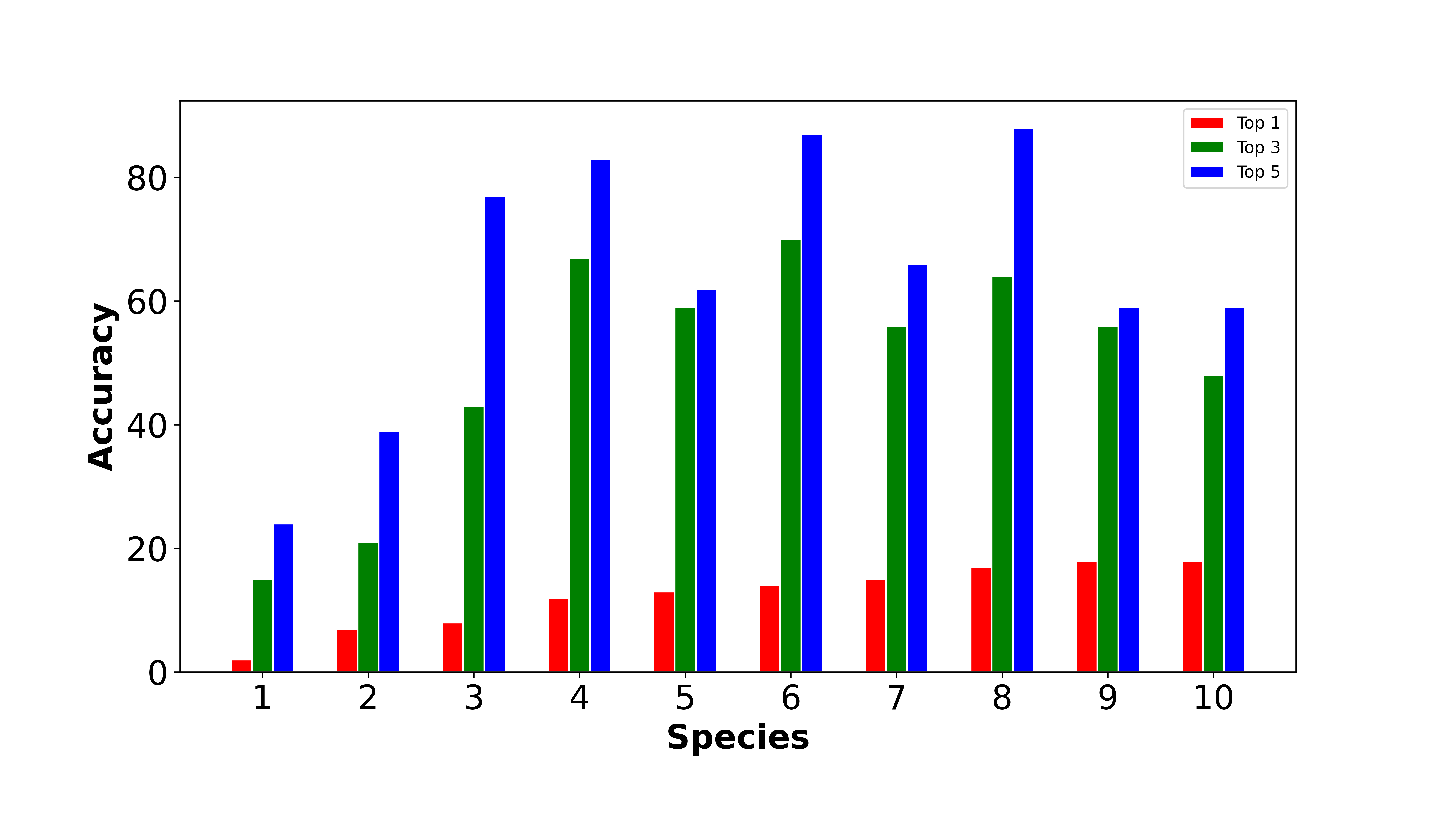}
\caption{The performance of ResNet-50 with the top 10 classes mentioned in Table \ref{tab:class_name_IP102}  in the dataset IP102.}
\label{fig:fig10a_ResNet_IP102}       % Give a unique label
\end{figure}

\begin{figure}[t]
\centering
\includegraphics[width=1.0\textwidth]{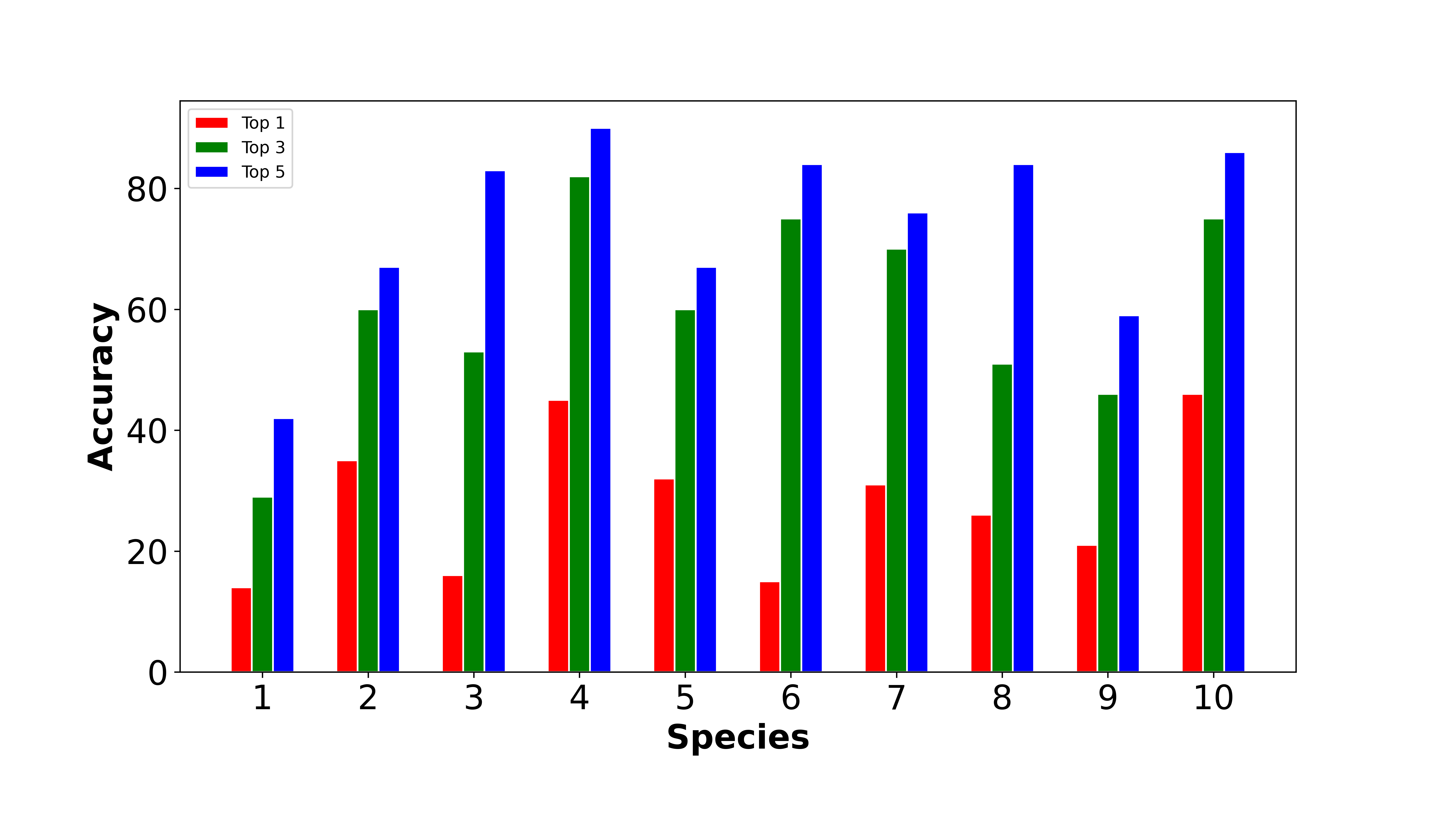}
\caption{The performance of MMAL-Net with the top 10 classes mentioned in Table \ref{tab:class_name_IP102}  in the dataset IP102.}
\label{fig:fig10b_MMAL_IP102}       % Give a unique label
\end{figure}

\begin{figure}[t]
\centering
\includegraphics[width=1.0\textwidth]{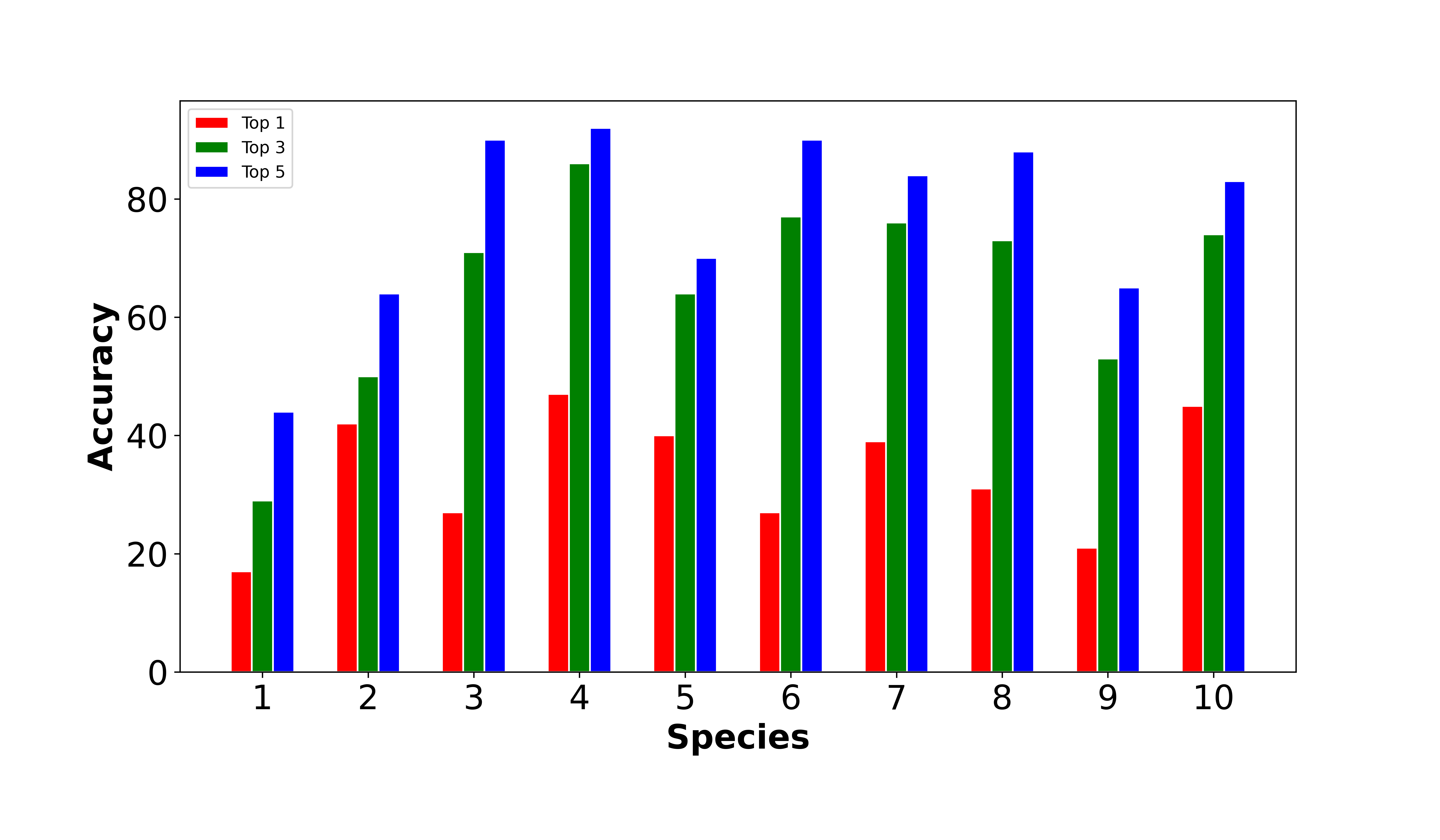}
\caption{The performance of our proposed ensemble method with the top 10 classes mentioned in Table \ref{tab:class_name_IP102}  in the dataset IP102.}
\label{fig:fig10c_Ensemble_IP102}       % Give a unique label
\end{figure}

\begin{figure}[t]
\centering
\includegraphics[width=1.0\textwidth]{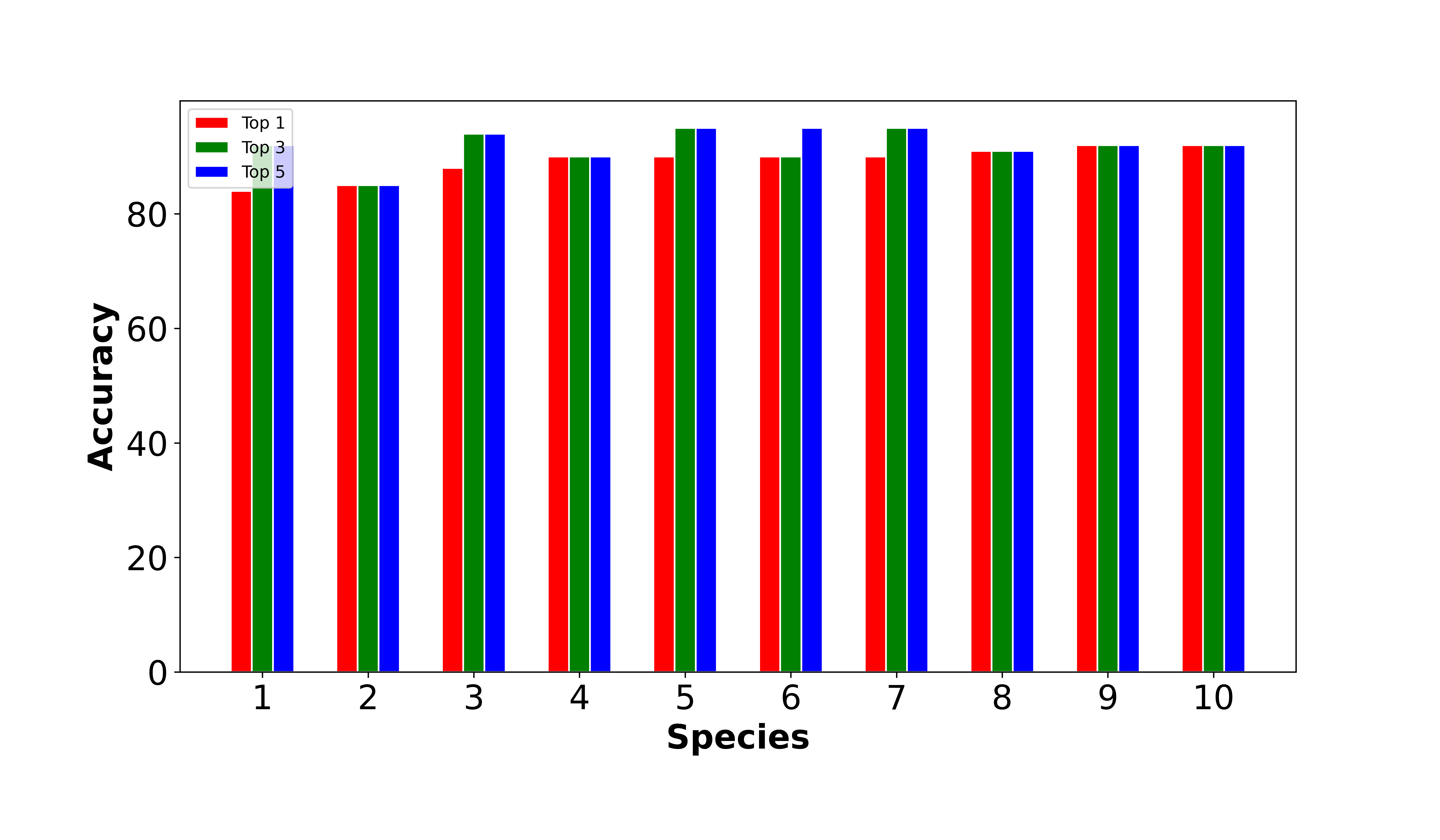}
\caption{The performance of ResNet-50 with the top 10 classes mentioned in Table \ref{tab:class_name_D0}  in the dataset D0.}
\label{fig:fig11a_ResNet_D0}       % Give a unique label
\end{figure}

\begin{figure}[t]
\centering
\includegraphics[width=1.0\textwidth]{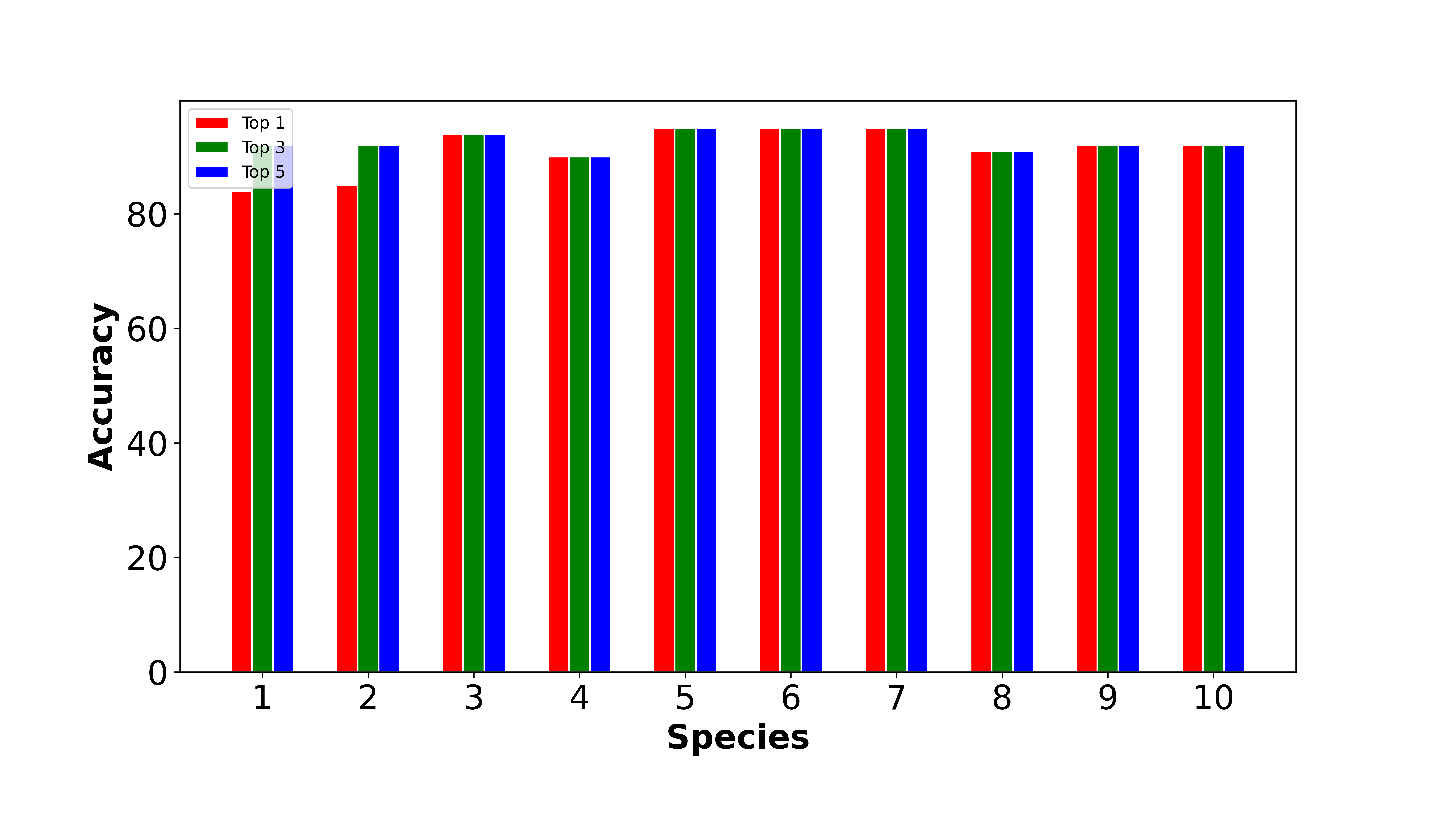}
\caption{The performance of MMAL-Net with the top 10 classes mentioned in Table \ref{tab:class_name_D0}  in the dataset D0.}
\label{fig:fig11b_MMAL_D0}       % Give a unique label
\end{figure}

\begin{figure}[t]
\centering
\includegraphics[width=1.0\textwidth]{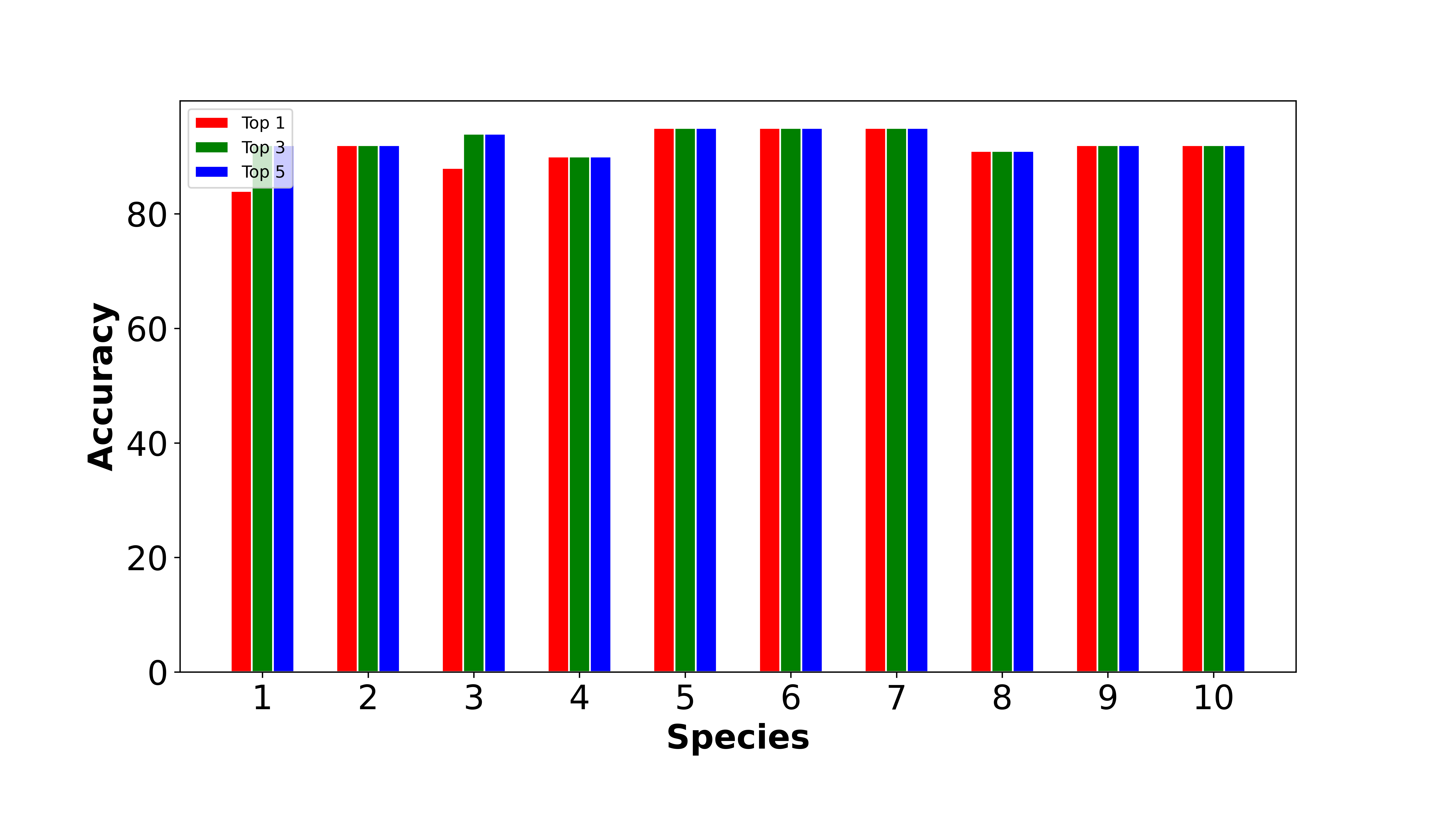}
\caption{The performance of our proposed ensemble method with the top 10 classes mentioned in Table \ref{tab:class_name_D0}  in the dataset D0.}
\label{fig:fig11c_EM_D0}       % Give a unique label
\end{figure}

Tables \ref{tab:class_name_IP102} and \ref{tab:class_name_D0} show the list of top 10 classes having the lowest accuracy using ResNet-50 on the dataset IP102 and D0, respectively. As visualized in Figs. \ref{fig:fig10a_ResNet_IP102}, \ref{fig:fig10b_MMAL_IP102} , and \ref{fig:fig10c_Ensemble_IP102} , the performance of MMAL-Net and our ensemble method is much better than the corresponding performance using ResNet-50 for all top 1, 3, and 5 accuracy in these two datasets. Between MMAL-Net and the proposed ensemble method, our ensemble model could achieve a better performance in these 10 classes in the dataset IP102.
One can see similar behaviors in the dataset D0, as depicted in Figs. \ref{fig:fig11a_ResNet_D0}, \ref{fig:fig11b_MMAL_D0}, and \ref{fig:fig11c_EM_D0}.

\begin{table}
\centering
\caption{The top 10 classes having the lowest accuracy using ResNet-50 on the dataset IP102. }
\label{tab:class_name_IP102}         % Give a unique label
\begin{tabular}{llll}
\hline\noalign{\smallskip}
Number & Species & Accuracy(\%) \\
&&from ResNet50\\
\noalign{\smallskip}\hline\noalign{\smallskip}
1 & Therioaphis maculata Buckton & 16.67 \\
2 & Beet fly & 25.0 \\
3 & Green bug & 25.51\\
4 & Polyphagotars onemus latus & 26.92\\
5& Large cutworm & 27.03\\
6 & Rice shell pest & 27.64\\
7& English grain aphid & 32.49\\
8 & white margined moth & 33.33\\
9 & Bird cherry-oataphid & 34.27\\
10 & Mango flat beak leafhopper & 35.71\\
\noalign{\smallskip}\hline
\end{tabular}
\end{table}

\begin{table}
\centering
\caption{The top 10 classes having the lowest accuracy using ResNet-50 on the dataset D0. }
\label{tab:class_name_D0}       % Give a unique label
\begin{tabular}{llll}
\hline\noalign{\smallskip}
Number & Species & Accuracy(\%)\\
&& from ResNet50 \\
\noalign{\smallskip}\hline\noalign{\smallskip}
1 & Nilaparvata lugens Stl & 84.62\\
2 & Pieris rapae Linnaeus& 85.71\\
3 & Dolycoris baccarum Linnaeus&88.24\\
4& Dryocosmus KuriphilusYasumatsu & 90.00\\
5& Halyomorpha halys Stl & 90.00\\
6 & Luperomorpha suturalis Chen&90.00\\
7& Riptortus pedestris Fabricius &90.91\\
8 & Laodelphax striatellus Falln &91.67\\
9 & Chauliops fallax Scott &92.31\\
10 & Plutella xylostella Linnaeus &92.31\\
\noalign{\smallskip}\hline
\end{tabular}
\end{table}

\subsection{Visualization with Grad-CAM}
This section presents visualization for our proposed models to show where those models focus on the input image to make the predictions. We utilized the Gradient-weighted Class Activation Mapping (Grad-CAM) proposed in~\cite{DBLP:journals/corr/SelvarajuDVCPB16}. In object classification, Grad-CAM commonly uses the computed gradient of a given target class flowing through the final convolutional layer of the feature extractor part to produce class activation maps (CAMs). Our paper visualized CAMs using the gradient flowing through the last feature extractor block of each proposed model.

\begin{figure}[t]
%\captionsetup{justification=centering}
\centering
% figures can also be resized by adding to the include graphics command as follows
\includegraphics[width=1.0\textwidth]{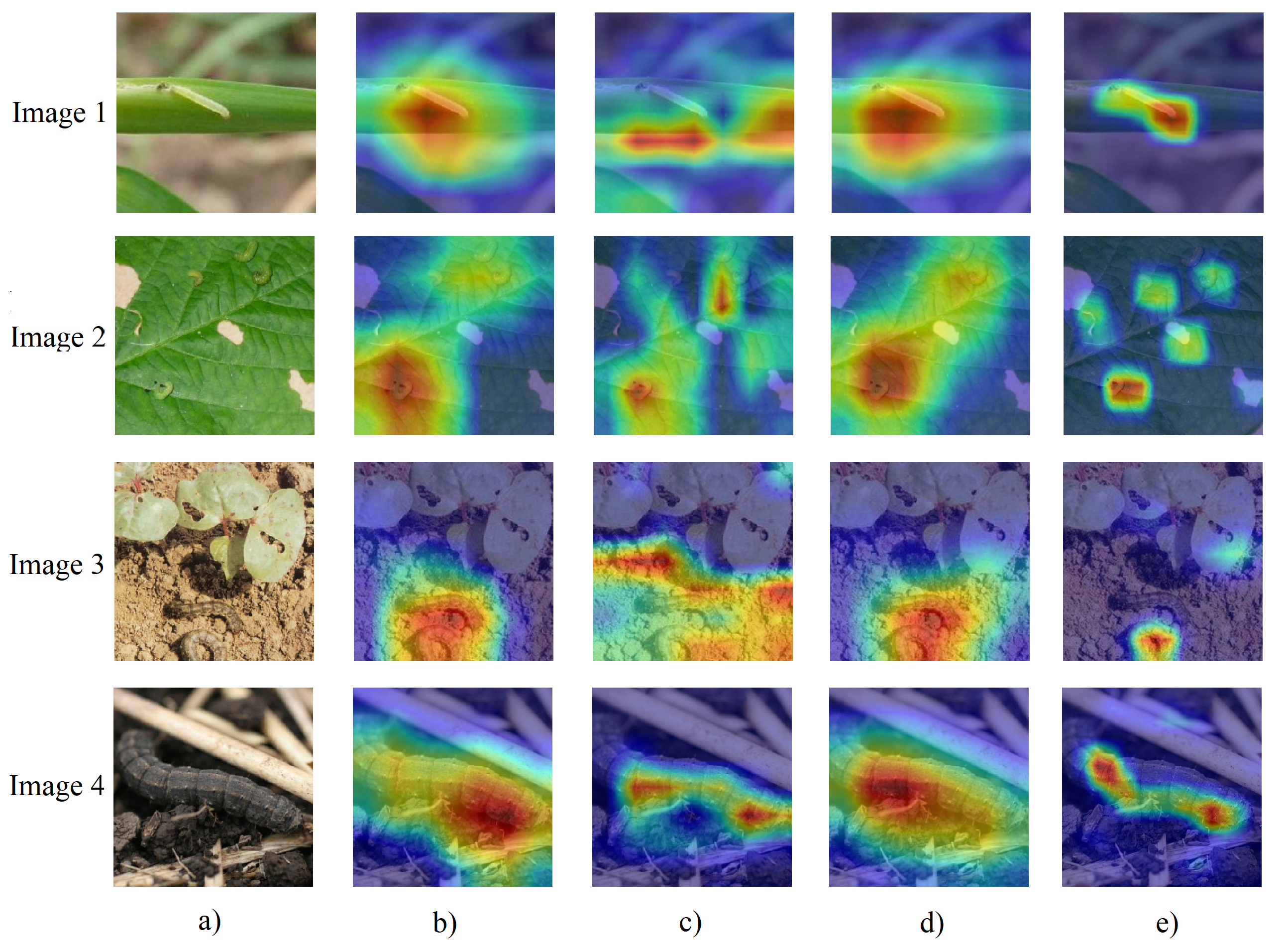}
\caption{Visualization of Grad-CAMs produced by ResNet-50 and our proposed models.
With the input images of IP102 in column (a), Grad-CAMs of ResNet-50 (column (b)), RAN (column (c)),  FPN (column (d)) and MMAL-Net (column (e)) are presented.}
\label{grad}       % Give a unique label
\end{figure}

Fig.~\ref{grad} shows Grad-CAMs produced by ResNet-50, RAN, FPN, and MMAL-Net using the input images in IP102 and their correct classes. They show that MMAL-Net performs the best on focusing the insects in the input images even those insects are small such as in image 2. ResNet-50 and FPN seem to perform the same, which correctly focus on the region containing the insects. On the other hand, RAN seems to focus on large and less accurate areas to make predictions.

\section{Conclusion and Future Works}
In this paper, we have investigated different CNN-based methods, i.e., Residual-Attention Network (RAN), Feature Pyramid Network (FPN), and multi-branch and multi-scale attention network (MMAL-Net) for recognizing insect pests. Among these methods, MMAL-Net can achieve the best accuracy of $72.15\%$ and $99.56\%$ on two datasets IP102 and D0, respectively. Furthermore, we visually validated that our models focused on the correct region, even on the input images of low-scale insects or a noisy background. With the combination of chosen models by the ensemble technique, we can obtain the better accuracy of $74.13\%$ and $99.78\%$ on IP102 and D0, respectively, and bypass the state-of-the-art methods related to the insect pest classification problem on these two datasets. For contributing to the research community, we publish all source codes associated with this work at \url{https://github.com/hieuung/Improving-Insect-Pest-Recognition-by-EnsemblingMultiple-Convolutional-Neural-Network-basedModels}. 

We aim to consider utilizing variants of CNNs to solve challenges in insect pest classification and applying efficient data augmentation methods in future work.

%BibTeX users please use one of
\nocite{*}
\bibliographystyle{spbasic}      % basic style, author-year 
\bibliography{reference}   % name your BibTeX data base

\end{document}